# Zero-Shot Digital Rock Image Segmentation with a Fine-Tuned Segment Anything Model

**Authors：**

Zhaoyang Ma[a]; Xupeng He[b]; Shuyu Sun[a*]; Bicheng Yan[a*]; Hyung Kwak[b]; Jun Gao[b]

[a]Earth Science and Engineering, Physical Science and Engineering Division, King Abdullah University of Science and Technology (KAUST)

[b]Saudi Aramco EXPEC Advanced Research Center, Reservoir Engineering Technology Division, Pore Scale Physics Focus Area, Bldg. 2291, GA-168

**Abstract：**

Accurate image segmentation is crucial in reservoir modelling and material characterization, enhancing oil and gas extraction efficiency through detailed reservoir models. This precision offers insights into rock properties, advancing digital rock physics understanding. However, creating pixel-level annotations for complex CT and SEM rock images is challenging due to their size and low contrast, lengthening analysis time. This has spurred interest in advanced semi-supervised and unsupervised segmentation techniques in digital rock image analysis, promising more efficient, accurate, and less labour-intensive methods.

Meta AI's Segment Anything Model (SAM) revolutionized image segmentation in 2023, offering interactive and automated segmentation with zero-shot capabilities, essential for digital rock physics with limited training data and complex image features. Despite its advanced features, SAM struggles with rock CT/SEM images due to their absence in its training set and the low-contrast nature of grayscale images. Our research fine-tunes SAM for rock CT/SEM image segmentation, optimizing parameters and handling large-scale images to improve accuracy. Experiments on rock CT and SEM images show that fine-tuning significantly enhances SAM's performance, enabling high-quality mask generation in digital rock image analysis. Our results demonstrate the feasibility and effectiveness of the fine-tuned SAM model (RockSAM) for rock images, offering segmentation without extensive training or complex labelling.

**Keywords**: Segment Anything Model (SAM); Digital Rock Physics; CT; Segmentation; Foundation Model; Zero-shot Generalization; SEM; RockSAM; Deep learning

# 1. Introduction

The recent advancements in semantic segmentation heavily rely on two key factors: end-to-end training of convolutional networks, and the availability of large-scale segmentation annotations. However, it's essential to underscore the challenging nature of obtaining such labeled data, especially within the digital rock community. Even though semantic segmentation models are making rapid progress, it remains evident that having access to large-scale training data significantly enhances accuracy, as indicated by findings in [1]. Yet, the process of annotating precise, mask-level labels for extensive image datasets is an arduous undertaking. Even with the aid of annotating tools, it still takes minutes for an experienced annotator to label a single image. This time-consuming task presents a significant challenge and may ultimately limit the amount of data that can be equipped with mask-level labels [2].

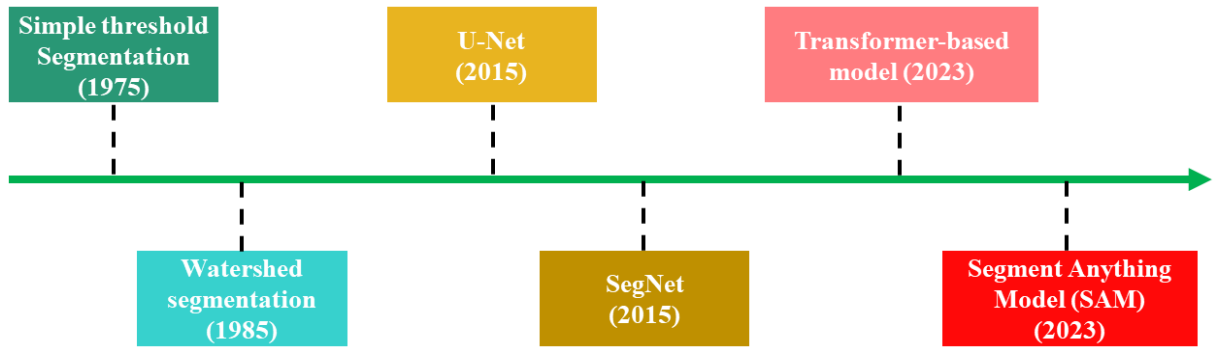

Fig. 1 Evolution of representative segmentation methods in the digital rock community.

Figure 1 highlights the key developments in image segmentation for digital rock physics. Initially, simple threshold segmentation [3] was employed, relying on pixel intensity. This method, known for its speed and simplicity, often falls short in handling digital rock images that contain excessive noise and artifacts. The watershed segmentation method [4] gained popularity in this domain due to its morphological approach, which focuses on topography to distinguish touching objects—a common occurrence in rock images comprising multi-minerals or grains. However, this technique's sensitivity to noise necessitates pre-processing with denoising algorithms. Beyond the influence of noise and artifacts, the resolution of rock images plays a crucial role in determining segmentation accuracy. A common trade-off exists between the field-of-view (FoV) and resolution, where achieving higher resolution frequently requires the use of smaller samples. To address this issue, various super-resolution techniques have been employed to enhance the resolution of CT images. Prominent among these methods are the Generative Adversarial Network (GAN) [5, 6] and diffusion models [7], which have demonstrated effectiveness in this context.

The evolution of machine learning and deep learning, alongside advancements in computational power and the availability of large image datasets, has led to the adoption of various deep learning methods. Given that traditional methods such as thresholding and watershed segmentation are prone to user bias [8, 9], the importance of adopting an automated segmentation technique for digital rock image analysis cannot be overstated. This approach, requiring minimal human intervention, is essential to ensure objective, consistent, and unbiased results in digital rock image segmentation. A significant breakthrough in 2015 was the introduction of U-Net [10], a convolutional neural network designed for biomedical image segmentation. U-Net stands out because it requires fewer training images yet achieves more accurate segmentation. This precision is crucial for the digital rock community, particularly in analysing porosity connectivity, which directly impacts the accuracy of fluid flow simulations in CT imaging [9].

U-Net has a symmetric U-shaped architecture, which consists of a contracting path (encoder) to capture the context and a symmetric expanding path (decoder) to enable the precise localization. Subsequently, several adaptations of U-Net, including U-Net++ [11], attention U-Net, and ResNet-U-Net [9], have been implemented [1]. SegNet [12], with its encoder-decoder architecture, stands out in efficiently handling large-scale segmentation tasks. Its distinctive feature lies in utilizing pooling indices within the decoder, employing memorized indices from the pooling layers during downsampling to upsample feature maps without additional learning. While SegNet might not achieve the high-precision segmentation of U-Net, especially in complex scenarios where fine details are essential, its efficiency in processing larger images (such as SEM images) makes it a viable option for digital rock physics applications [13].

Although U-Net and SegNet have gained widespread recent usage for their outstanding segmentation performance, leveraging the encoder-decoder structure, their effectiveness is somewhat limited by the inherent locality of convolution operations, which restricts their ability to model global context and long-range spatial dependencies [1, 14]. In contrast, Transformers, initially successful in natural language processing, have recently been adopted in vision tasks. Their strength lies in effectively modelling global contexts, a capability crucial for these applications. The current research acknowledges that pure transformer backbones have surpassed their CNN counterparts in image segmentation [15]. Notably, transformer-based models like TransUNet [1] have demonstrated superior performance over U-Net and attention-U-Net. However, these methods still fall under supervised learning as they rely on paired data (original digital rock images and their corresponding segmentation images). For example, in binary segmentation of CT images, an original image corresponds to a binary-segmentation image containing pores and rock matrix. This process is time-consuming and requires expert knowledge, not to mention the more complex multi-phase segmentation from thin-section images [16], where accurately identifying mineral types demands extensive expertise and time. Despite Ma et al. [1] utilized image augmentation techniques, the necessity for some pre-existing labeled data persists, owing to the inherently supervised learning nature of the U-Net model.

The limitations of most digital rock image segmentation models are significant, particularly their inability to process unstructured and unevenly distributed 3D data. Additionally, many existing models struggle to deliver real-time performance and are limited in their capacity to handle noise and variations in image quality. These challenges prompt the question: *is it possible to achieve segmentation without prior training (zero-shot generalization) and directly generate segmentation masks*? In 2023, Meta AI introduced the Segment Anything Model (SAM) [17], which enables zero-shot generation of segmentation masks. However, since SAM is primarily trained on natural images and not specifically on digital rock images (like CT, SEM, thin-section images), its performance on the latter is suboptimal and not directly transferable. Therefore, in this paper, we aim to fine-tune the SAM model to enhance its suitability for segmenting digital rock images, addressing the common limitations in current models related to real-time performance.

## 2. Methodology

### 2.1 Brief introduction of SAM

In the field of digital rock physics, segmentation methods are primarily split into two distinct approaches. The first and more prevalent method is automatic segmentation, which is tasked with identifying specific categories within images, such as differentiating between pores and rock matrices in binary segmentation, or discerning rock-forming minerals and fractures in multi-phase segmentation. This approach relies heavily on large collections of manually annotated examples, which may number in the thousands, and requires significant computational resources and expertise for model training. However, the process is time-intensive and prone to human error, particularly in the precise classification of complex features such as rock-forming minerals, which demands expert knowledge and high-quality, noise-free image data. The second is interactive segmentation, involving experts who manually adjust masks to segment objects in digital rock imagery. Both approaches, however, are not fully automated and present limitations in segmentation tasks.

In contrast, the segment anything model (SAM) model [17] represents an advancement by combining the strengths of both interactive and automatic segmentation. Often likened to a pivotal moment similar to ChatGPT's impact on the AI community, SAM has been a buzzword recently and it stands as the foundational model for image segmentation. It operates through a flexible, promptable interface, enabling a wide array of segmentation tasks by simply crafting the appropriate prompts, such as clicks or text descriptions. Trained on a vast and varied dataset comprising over 1 billion masks, SAM boasts remarkable generalization capabilities. It can recognize and segment new object types and images that were not part of its training set. This generalization obviates the need for practitioners to gather bespoke segmentation data or to intricately adjust models for specific use cases.

The SAM [17] model, which is trained on the Segment Anything 1 Billion (SA-1B) dataset, which is the largest segmentation dataset so far, may not perform optimally on digital rock images due to the specificity of the dataset it was trained on. The SA-1B dataset comprises over 1.1 billion segmentation masks that are derived from approximately 11 million images. These images are diverse and privacy-conscious, tailored for the development of models capable of general object segmentation in a variety of open-world scenarios. However, digital rock images usually require specialized segmentation due to their unique textures, patterns, and geological features, which are not adequately represented in the SA-1B dataset. Consequently, the SAM model's training on a general and broad dataset might not equip it with the necessary nuances to accurately segment the specialized features present in digital rock imagery [18].

We adopted the SAM model for the segmentation of digital rock images because it offers four distinct functionalities that are particularly advantageous:

1. It streamlines the segmentation process in digital rock physics by allowing users to quickly select objects with a single click or through an interactive process that involves marking points to be included in or excluded from the object's boundary. The segmentation can also be initiated by drawing a bounding box around the object of interest, which is useful for isolating specific rock features like mineral grains or void spaces.

2. In cases where it's challenging to determine the exact feature to segment, such as differentiating between closely packed mineral grains, SAM has the ability to generate multiple valid masks. This adaptability is key for addressing the ambiguities often encountered in digital rock physics imaging.

3. SAM has the autonomous ability to detect and segment all visible features within a digital rock image, which is essential when dealing with complex images where manual identification of every feature is impractical.

4. Once the image's embedding has been precomputed, SAM is capable of producing segmentation masks instantly in response to any query. This allows for real-time segmentation, a significant advantage when rapid analysis of rock features is required, such as in dynamic porosity evaluations or real-time microstructure analysis.

Nevertheless, the pre-trained SAM model [17] developed by Meta AI exhibits two significant limitations. Firstly, it often produces coarse mask boundaries, which may overlook the segmentation of thin or small object structures, leading to incorrect predictions or substantial errors [18]. This drawback can notably limit the applicability and effectiveness of SAM in automated annotation tasks. Such tasks, especially in fields like digital rock physics, demand highly accurate image masks for precise analysis and interpretation. Secondly, it cannot use for the digital rock images because the low-contrast as well as different imaging mechanism for X-ray CT data or SEM images with the natural images.

The limited use of the pre-trained SAM model in processing rock images can be attributed to several key factors: Firstly, rock images, especially those acquired through techniques like Scanning Electron Microscopy (SEM) or CT scans, exhibit unique characteristics. These include intricate patterns, diverse textures, and subtle contrasts, which starkly contrast the natural images that SAM was initially trained on. This difference poses a significant challenge for the model to segment rock images accurately without extensive retraining or fine-tuning. Moreover, even after fine-tuning, there's a possibility that the model may not effectively generalize across various types of rock images. Rock images can differ greatly in their physical and chemical properties, necessitating a model capable of adapting to a broad spectrum of features – a substantial hurdle in the realm of deep learning.

## 2.2 Mask-generation ability of SAM

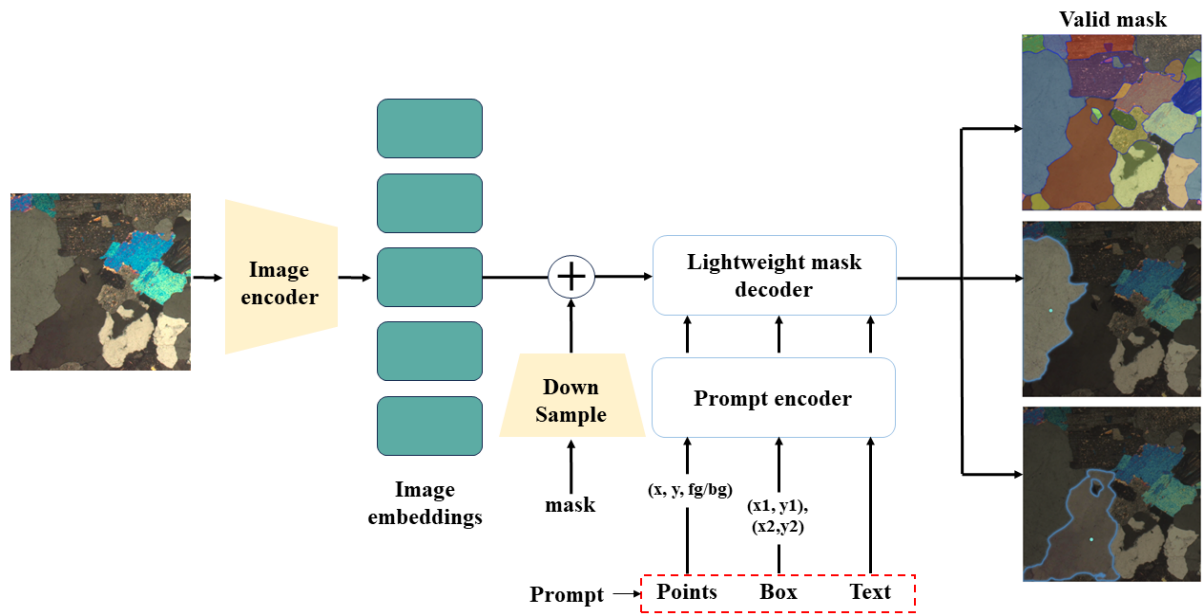

Fig. 2 Segment Anything Model (SAM) overview

Figure 2 presents a comprehensive depiction of the Segment Anything Model (SAM). This process begins by encoding the image into a high-dimensional vector. Concurrently, the provided prompt (points, a box, or text) undergoes encoding into its own distinct vector representation. These two vectors are then amalgamated and directed through a mask decoder [19]. The outcome of this is a mask tailored to the object delineated by the prompt. The image encoding component utilizes a Vision Transformer model, a substantial language model pre-trained on an extensive collection of images. Please be noted that a language model can be utilized for image analysis tasks by first encoding the image into a text representation. The prompt encoding is facilitated by a straightforward text encoder, transforming the input prompt into a vector format.

The significance of prompts in the Segment Anything Model (SAM) is multifaceted. Primarily, prompts offer precise guidance to the model, delineating the specific object or feature to be segmented in the image. This clarity is essential, as the absence of a distinct prompt could lead the model to misinterpret the focus of the image, resulting in imprecise or irrelevant segmentations. Furthermore, prompts enhance customization and flexibility in the segmentation process. The ability to utilize various prompt types, such as points, boxes, or text, allows for a tailored approach to segmentation. Text prompts, for example, enable abstract or conceptual specification of objects, while points and boxes provide concrete spatial direction. Additionally, clear and detailed prompts aid the model in generating more accurate probability maps, which assess the likelihood of each pixel belonging to a specific category or object. This clarity in guidance minimizes ambiguity and significantly improves the segmentation's accuracy.

Finally, the mask decoder, a nimble transformer model, deduces the object mask based on the combined embeddings of the image and prompt.

### 2.3 Why fine-tune a model?

The SAM model exhibits remarkable zero-shot generalization capabilities, effectively adapting to new data across varied distributions and tasks, as noted in reference [20]. However, its performance is less effective in certain domains, including medical image segmentation and digital rock imagery, as indicated by findings in reference [21]. This limitation underscores the necessity for additional refinement or the exploration of alternative methodologies within these specialized domains. Consequently, fine-tuning the SAM model specifically with digital rock images emerges as a more viable and promising strategy. Such tailored adjustment could significantly enhance the model's performance in segmentation tasks, particularly in practical rock engineering applications, where precise and efficient image analysis is paramount.

Fine-tuning a model entail adapting a pre-trained model to excel in a new, specific task or dataset, thereby enhancing its performance on novel data. In the context of digital rock image segmentation, fine-tuning the SAM model differs significantly from initiating training from scratch due to the initial values of the weights and biases. When training begins from the scratch, weights and biases are randomly assigned based on specific strategies. This initial setup means the model lacks any preliminary knowledge about the task, often leading to suboptimal performance. Conversely, starting with pre-established weights and biases leverages existing knowledge, enabling us to refine these parameters more effectively for our specialized dataset. For instance, the skills developed for identifying certain features in one type of image can be adaptively applied to similar tasks, such as recognizing distinct but related patterns or structures in rock images. This approach of fine-tuning enhances the model's ability to discern and segment intricate details in digital rock imagery, capitalizing on previously learned information.

Fine-tuning the SAM model for digital rock images involves modifying the weights in the mask decoder (Fig.2), while maintaining the other components as they are. To reduce computational demands, the image encoder is kept static, as it accounts for the majority of the computational load in SAM. The prompt encoder, which encodes the positional information of the bounding box, can be reused from SAM's pre-trained bounding-box encoder, so this component is also left unchanged [19]. The remaining section is the fine-tuning mask decoder, this process necessitates digital rock images along with their corresponding masks, as shown in Figure 3. The CT images and labeled data used pertain to Leopard sandstone, sourced from the Digital Rock Portal (https://www.digitalrocksportal.org/projects/317). The binary image data, derived from filtered grayscale images, was segmented at a threshold level of 72, determined using the IsoData algorithm. For further information, please refer to [22].

For the prompt component—be it points, boxes, or text—the bounding boxes of each object are calculated for use as prompts. As previously mentioned, effective prompts are crucial for the model's learning efficiency. They offer clear, consistent training signals, enabling the model to more effectively learn relevant features and patterns. In complex scenes with multiple objects, prompts are invaluable for distinguishing between different elements, particularly when segmenting a specific object from multiple options. Moreover, well-crafted prompts can reduce computational demands by directing the model's focus more precisely to the area of interest, instead of uniformly processing the entire image.

Fine-tuning a model offers several benefits: it enhances the model's performance, conserves computational resources, and reduces training costs. This process allows the model to utilize pre-trained knowledge for the target task, enabling adaptation to previously unseen or new data distributions. Additionally, fine-tuning tailors the model to specific use cases, thereby optimizing its performance for those particular scenarios.

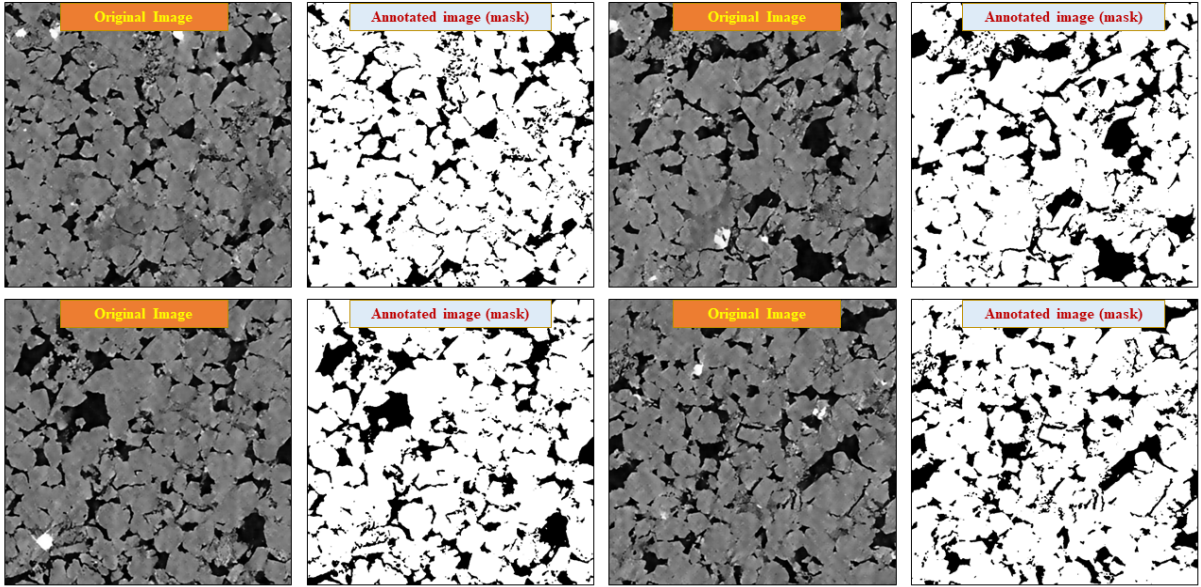

Fig. 3 Representative training dataset for fine-tuning the SAM Model: Leopard sandstone images (400 images each with 1000×1000 in size)

In the ensuing section of our paper, we present a comprehensive breakdown of the process to fine-tune the Segment Anything Model (SAM) specifically for the task of segmenting digital rock images. This detailed exposition is based on the Python code provided and unfolds through several critical stages, each with distinct components.

## 2.4 Implementation details

### 2.4.1 Environment Setup and Data Preparation

1. Importing Essential Libraries: Our methodology is fundamentally grounded in importing critical Python libraries, including OpenCV, NumPy, PyTorch, and Transformers. The integration of these libraries plays a crucial role in enabling a spectrum of operations such as image processing, data manipulation, and various deep learning functions, thereby laying the groundwork for efficient model handling and manipulation. Of particular importance is the adoption of Patchify, a key tool in our process. Given the typically large size of CT images of rocks, often around 1000×1000 or 2000×2000 pixels, Patchify becomes indispensable. It allows us to efficiently divide these large images into smaller, regular grid patches with the size of 256×256, which are more manageable for training. This step is vital as it not only makes the processing of large images feasible but also ensures uniformity and consistency in the data fed to the model, significantly impacting the overall effectiveness of our segmentation approach.

2. Configuring Constants and Parameters: Following the library importation, we establish a well-defined configuration. This phase involves meticulously setting paths for image and mask directories, along with defining key parameters such as patch size, batch size, learning rate, and other vital hyperparameters. Such a configuration underpins a consistent and controlled training environment, essential for the reproducibility and reliability of our results.

3. Initializing Distributed Training: Recognizing the computational demands of our task, we design the training process for a distributed environment. This approach is fundamental in efficiently managing large datasets, making optimal use of multiple GPUs. Key functions like 'setup_distributed()' and 'destroy_distributed()' are integral in managing the initialization and subsequent clean-up of this distributed setup.

### 2.4.2 Dataset Processing

1. Bounding Box Extraction: In this phase, the 'get_bounding_box()' function is employed to compute bounding boxes from ground truth masks. These boxes are pivotal in directing the model's focus to pertinent regions within the images, thereby enhancing the precision of segmentation.

2. Dataset Splitting and Management: We then split our dataset into training and validation subsets using the 'split_dataset()' function. This strategic division is crucial in striking a balance between effective learning and comprehensive model validation.

3. Patch Extraction and Normalization: Our methodology further involves processing images through the 'load_data()' function. This function is designed to segment large images into smaller, manageable patches and normalize the masks, ensuring the selection of patches with relevant and meaningful data for training.

### 2.4.3 Model Configuration and Fine-tuning

1. Initial Setup of Model and Processor: The SAM model, along with its processor, is initialized leveraging pre-trained weights. This critical step allows us to utilize the pre-existing knowledge embedded in the model, adapting it specifically for the nuanced task of digital rock image segmentation.

2. Preparation of Dataset and DataLoader: We create a custom 'SAMDataset' class tailored to the specifics of our dataset. Correspondingly, DataLoaders for both training and validation datasets are prepared, facilitating efficient batch processing during the training phase.

3. Fine-Tuning the SAM Model: A focused approach is taken in fine-tuning the SAM model, specifically targeting the mask decoder. Simultaneously, we freeze the vision and prompt encoders. This strategy concentrates the learning process on the segmentation task, while retaining the valuable pre-trained knowledge of the model.

### 2.4.4 Training Process

1. Loss Function and Optimizer Configuration: For the training process, we employ the 'DiceCELoss' function in combination with the Adam optimizer [23]. This choice is particularly effective for segmentation tasks, striking a balance between Dice loss (for assessing overlap) and cross-entropy loss (for pixel-wise classification).

2. Conducting the Training Loop: The model undergoes a rigorous training regimen for a predefined number of epochs. During each epoch, the model is evaluated against the validation set, with learning rate adjustments being made through a 'ReduceLROnPlateau' scheduler. This scheduler refines the learning process based on the observed validation loss.

3. Model Evaluation and Preservation: Continuous monitoring of the model's performance is conducted, with improvements in validation loss prompting the saving of the model. This ensures that the most effective version of the model is retained. Additionally, an early stopping mechanism is implemented to prevent overfitting.

### 2.4.5 Post-Training Cleanup

Upon completion of the training, we undertake a thorough cleanup of the distributed environment and ensure the efficient release of GPU resources. This final step marks an efficient conclusion to the training process.

Through this detailed methodology, we elucidate the steps involved in adapting the SAM model to the specific task of digital rock image segmentation, with a particular focus on fine-tuning the model to enhance its performance in this specialized field.

## 3. Case studies and results

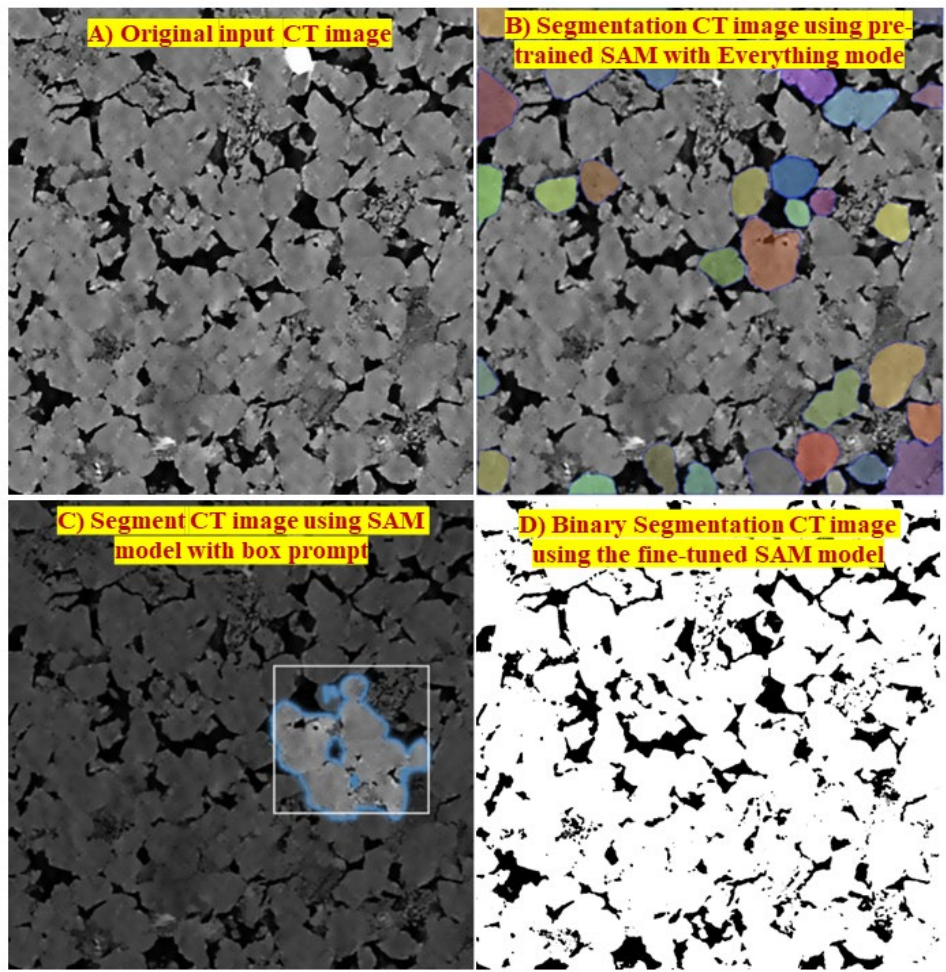

Fig. 4 Comparison between the segmentation mask generation by using the original SAM model with fine-tuned SAM model (RockSAM).

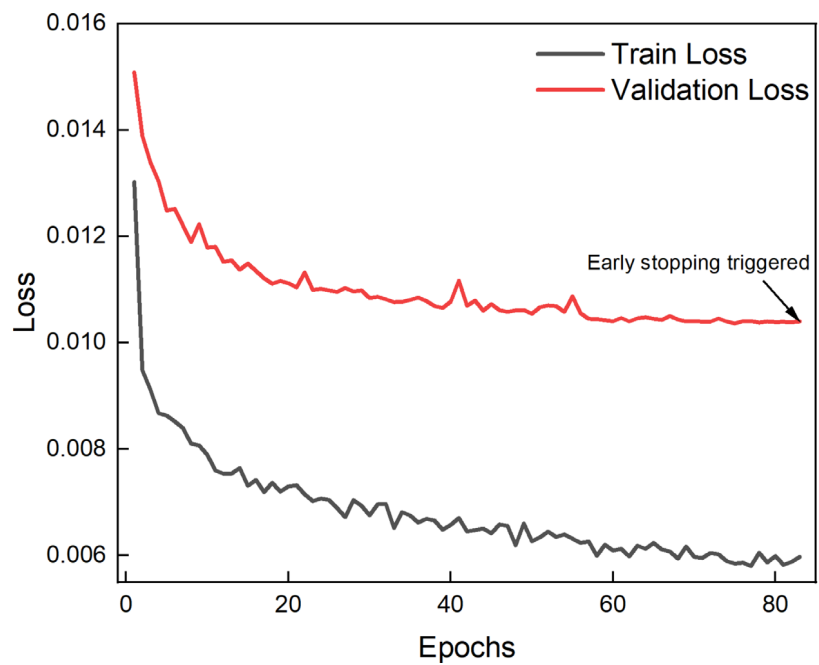

Fig. 5 Training and validation loss versus epochs during the training with only CT data.

As depicted in Figure 4, the Segment anything model (SAM) is innovatively designed for a novel, adaptable segmentation task. It enables zero-shot image segmentation using a pre-trained model through two primary modes: the automatic 'everything' mode and the manual 'prompt' mode, which includes bounding boxes, points, or textual prompts [24]. However, since SAM is primarily trained for natural images, it struggles to capture all major features in input digital rock images (CT image of rock), highlighting a limitation in its effectiveness. We compared different settings to fully explore the performance of SAM under various strategies. As shown in Fig. 4B, we first apply the 'everything' mode of SAM to CT images, it is observed that the model only extracts some features, missing most objects. After that, Fig. 4C demonstrates that we used a bounding box as a prompt for the segmentation of digital rock images, which results in apparent inaccuracies. Therefore, fine-tuning the SAM model is essential to achieve accurate segmentation masks for these specific image types (Fig. 4D).

The SAM model is grounded in the vision transformer architecture [25], which processes images in a way that mirrors how textual data is handled by transformers. Tailored from the foundational "facebook/sam-vit-base" model, this refined SAM version is specifically engineered to meet the intricate segmentation demands of digital rock imagery. We have configured the ViT-Base image encoder, updating a total of 6.32 million parameters in the process. Differing from numerous studies that rely on A100 GPUs, all our experiments are designed to be executable on more widely accessible GPUs, enhancing the practicality and accessibility of our research. Fig. 5 demonstrates the training and validation loss versus epochs during the training process. For the inference section, our approach begins with the loading of a pre-trained SAM model configuration and processor. This model is then fine-tuned further with weights from a saved model, tailored to the specific nuances of digital rock structures. The model operates in a parallelized manner on a GPU, ensuring efficient processing of large datasets.

The fine-tuned SAM model (RockSAM) outputs a probability map for each patch, indicating the likelihood of each pixel belonging to a specific class. This soft mask is converted into a hard mask using a thresholding technique, providing a clear segmentation of the rock structures. The framework also includes a visualization component. The segmented images, probability maps, and predictions are displayed side by side for easy comparison and analysis (Fig. 6). This visualization not only aids in interpreting the results but also provides an intuitive way to assess the model's performance.

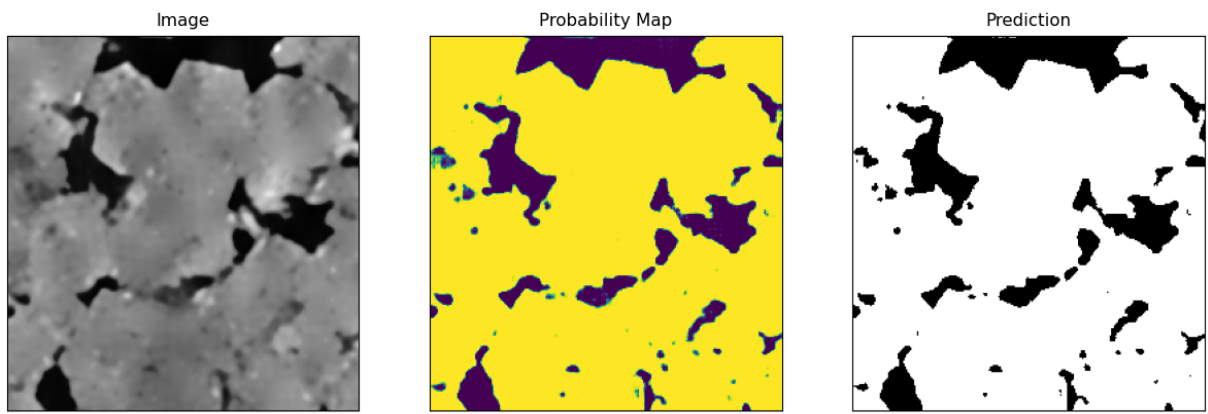

Fig. 6 A small patch of a large CT image with probability map and prediction of segmentation image

In conclusion, the fine-tuned SAM model offers a sophisticated and efficient tool for semantic segmentation in digital rock physics. Its ability to process large images in patches and its transformer-based architecture make it particularly suited for the intricate patterns and structures typical in geological formations. This paper delves into the specifics of the model's application in digital rock physics, demonstrating its efficacy and potential for advancing the field.

We also assess the effectiveness of the fine-tuned SAM to the SEM images of rocks, which is great difference with the training data, because all the training data are CT images of sandstone. Figs. 7-8 demonstrate that even for large-scale SEM images with dimensions of 4096x4096 pixels, satisfactory results can be achieved, albeit with occasional errors. These can be mitigated through the application of smooth blending techniques and employing a larger or more advanced version of the SAM model. In order to assess the segmentation performance of RockSAM, three commonly utilized metrics: Intersection over Union (IoU) [26], Dice similarity coefficient (Dice) [27] and mean absolute error have been adopted.

$$IoU\ (P,T) = \frac{P \cap T}{P \cup T} \quad (1)$$

The Intersection over Union (IoU) metric, denoted as $IoU\ (P,T)$, quantifies the overlap between the predicted segmentation mask, $P$, generated by RockSAM, and the ground truth mask, $T$. An IoU value of 1 signifies a flawless prediction, indicating a pixel-perfect alignment between the predicted segmentation mask and the ground truth.

$$Dice\ (P,T) = \frac{2P \cap T}{P + T} \quad (2)$$

$Dice\ (P,T)$ denotes the overlap between the predicted segmentation mask and the ground truth mask.

$$MAE(P,T) = \frac{\sum_{i=1}^{n} P_i - T_i}{N} \quad (3)$$

where $N$ is the number of pixels of an image, and $i$ is the $i$-th pixel of the image. A lower Mean Absolute Error (MAE) value signifies a more proficient segmentation model, characterized by enhanced accuracy in delineating precise segmentation boundaries and correctly identifying class labels.

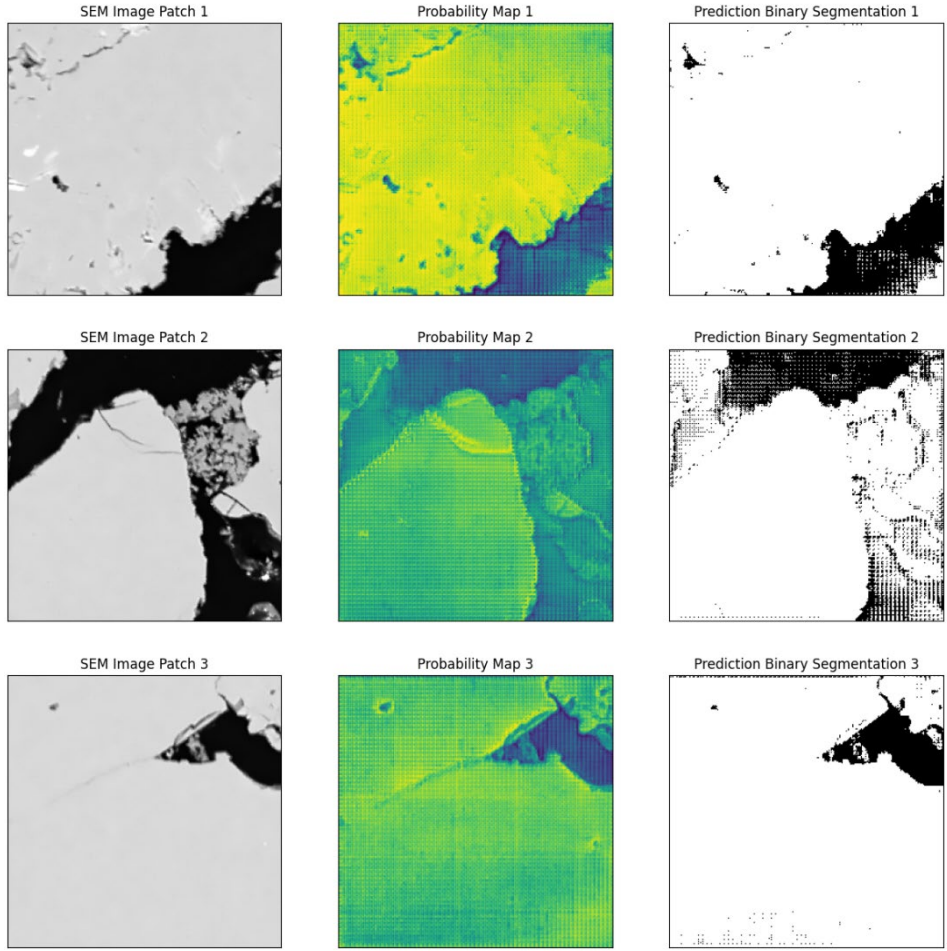

Fig. 7 Three distinct 256x256 patches from a large SEM image, showcasing corresponding probability maps and predictions of the binary segmentation images using the RockSAM model.

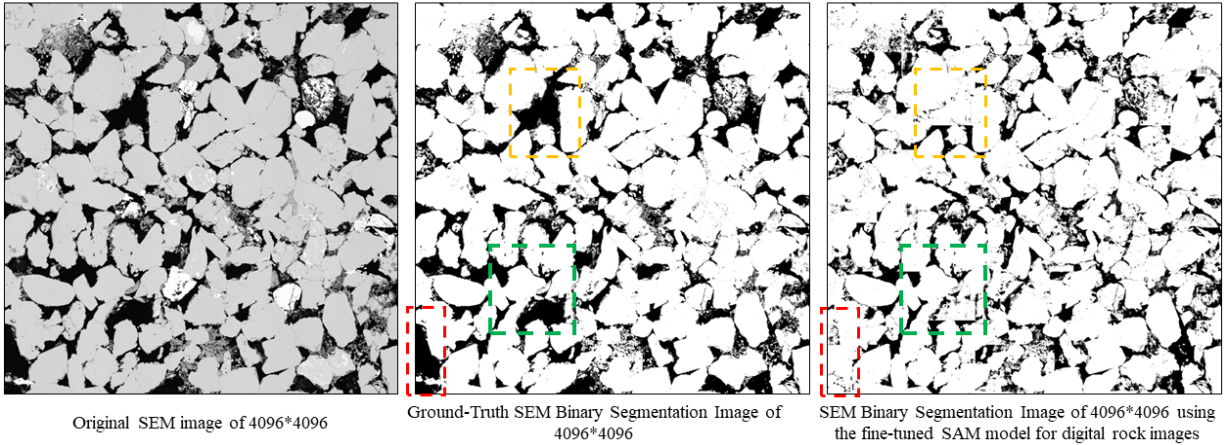

Fig. 8 Comparative analysis of ground truth SEM image and binary segmentation image generated by the fine-tuned SAM model with only CT data as training dataset.

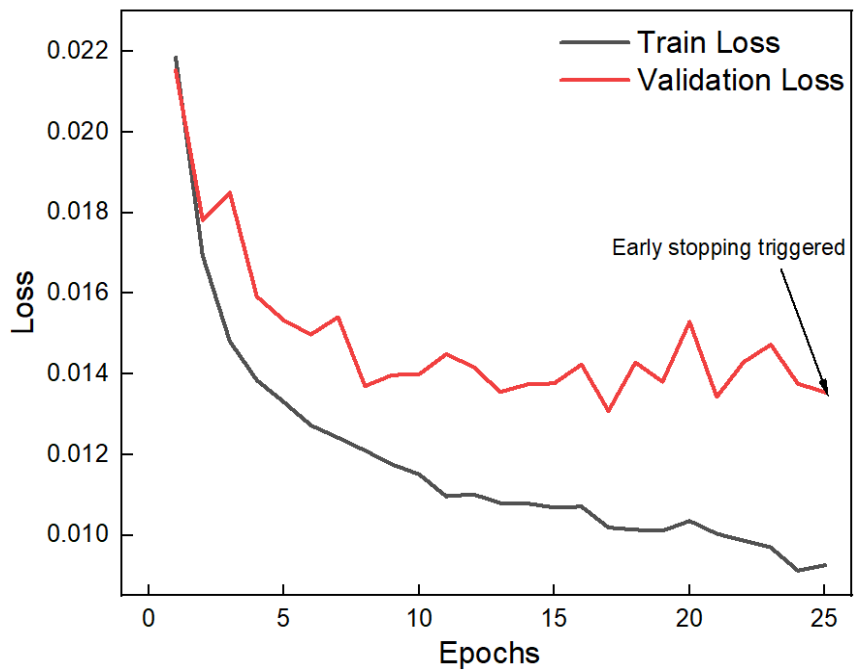

Fig. 9 Training and validation loss versus epochs during the training with both CT and SEM data.

Fig. 9 demonstrating the training and validation loss versus epochs during the training with both CT and data and figure 10 presents a thorough comparative analysis of binary segmentation on Scanning Electron Microscope (SEM) images of a digital rock, employing varying training datasets within the segmentation model. The figure is divided into several key sections, each highlighting different aspects and outcomes of the study:

1. Fig. 10A: This section showcases the original grayscale SEM image, vividly depicting the microstructure of the rock. SEM images are renowned for their high resolution and detailed portrayal of surface texture and composition. The intricate details visible in this image underscore the complexity of accurately segmenting such a nuanced structure.

2. Fig. 10B: Here, we observe the ground-truth binary segmentation image. This accurately segmented image is pivotal, serving as a benchmark for evaluating the performance of the segmentation model. The binary segmentation distinctly separates the image into two classes: pores and grains, providing a clear, unambiguous standard against which model outputs can be compared.

3. Fig. 10C: This part illustrates the binary segmentation image generated by a model solely trained on CT (computed tomography) data. CT scans, known for providing detailed 3D X-ray images, are adept at revealing internal structures. However, they might lack the comprehensive surface detail captured by SEM images. The segmentation accuracy here, when juxtaposed with the ground truth, appears less precise, discernible through the differences in highlighted areas.

4. Fig. 10D: In this segment, the binary segmentation image produced by the same model, but fine-tuned with an amalgamation of both CT and SEM data, is displayed. The rationale behind integrating both data types is compelling. SEM images contribute rich surface detail and resolution, elements that may be less pronounced in CT images. This synergy of surface details from SEM and internal structural insights from CT empowers the model to develop a more holistic representation of the rock's features. The resulting segmentation is markedly more accurate and detailed, aligning more closely with the ground truth presented in Fig. 10B.

The strategic inclusion of SEM images alongside CT data in the training dataset is a pivotal decision. It significantly enhances the model's comprehension of the rock's textural and morphological characteristics—factors that are essential for applications such as porosity analysis or material characterization. The nuanced detail afforded by SEM images enriches the model's ability to differentiate between various microstructural features, which might be less apparent or discernible in CT data alone.

A crucial aspect of this study is its demonstration of the potential to fine-tune the SAM in a rock context, as well as the observation that segmentation accuracy rapidly increases with the expansion of the dataset size. Currently, our dataset encompasses only 400 CT images and 59 SEM images, a relatively small collection compared to the 11 million images used in the training of the pre-trained SAM. This stark contrast in dataset sizes indicates a significant potential for improvement. It is anticipated that with the augmentation of the training dataset—both in terms of size and diversity (potentially incorporating other digital rock images, such as thin-section images, in the future)—the accuracy of segmentation will correspondingly escalate. This progression is not merely a theoretical expectation but a substantiated prediction based on the observed trends and outcomes within this study.

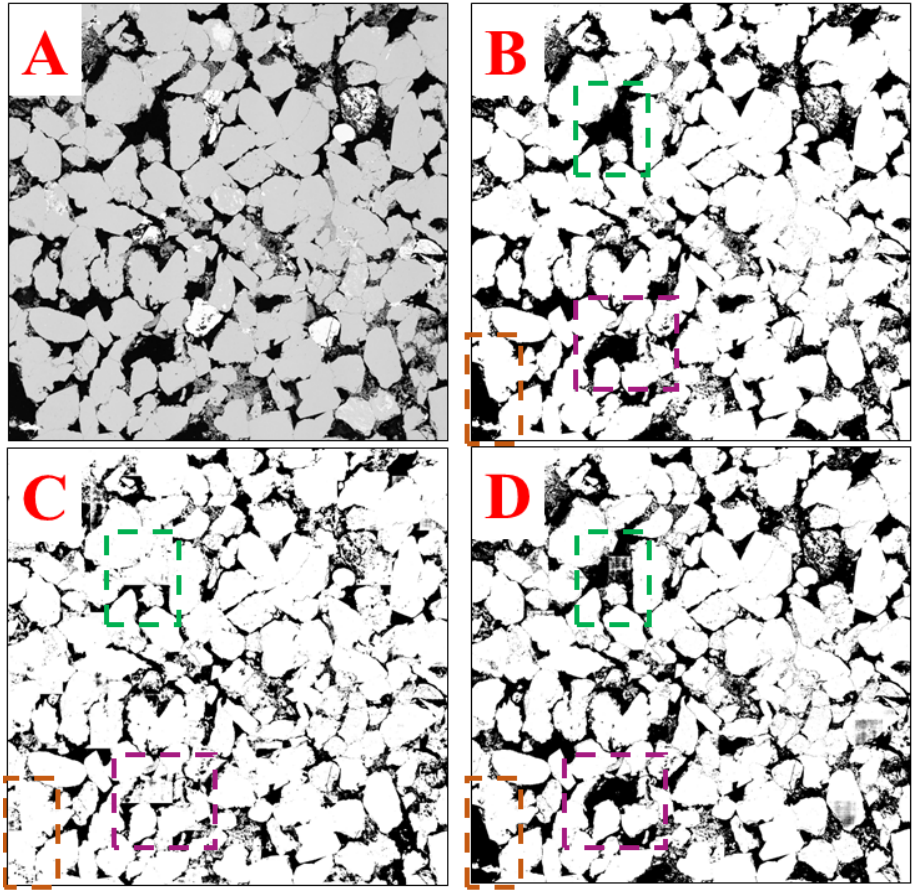

Fig. 10 Comparative analysis of ground truth SEM image and binary segmentation image generated by the fine-tuned SAM model with both CT and SEM data as training dataset. (A): Original SEM image of 4096*4096 (B): Ground-truth SEM binary segmentation image of 4096*4096 (C): SEM binary segmentation image using the fine-tuned SAM model for digital rock images with only CT data (D): SEM binary segmentation image using the fine-tuned SAM model for digital rock images with both CT and SEM data.

Figure 11 illustrates the application of the SAM model in identifying fractures in rock images, a feat that is quite remarkable. However, it's evident that this method primarily extracts larger fractures, while smaller ones are not detected as accurately. Therefore, further refinements are needed for employing the SAM model in the segmentation of digital rock images. Nonetheless, the SAM model in digital rock image is anticipated to gain substantial popularity in the coming years.

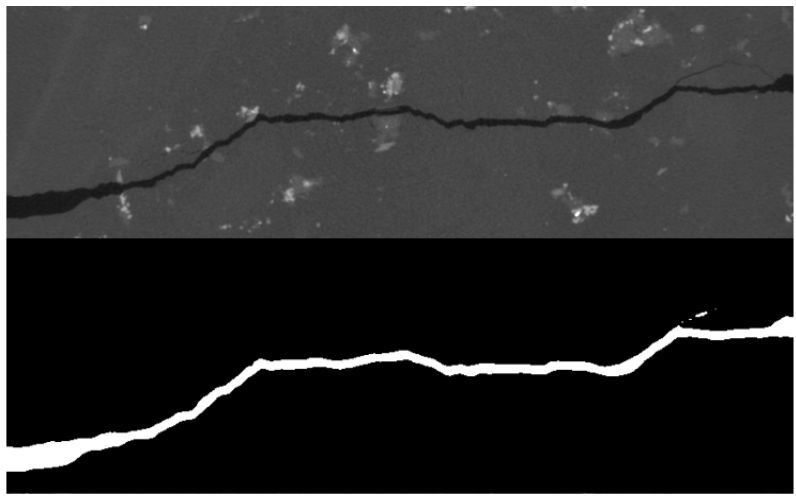

Fig. 11 Zero-shot segmentation of fractures using the SAM model.

## 4. Discussion: Limitations and Future work

The task of accurately segmenting digital rock images is fraught with challenges, primarily due to their complex and minute features, compounded by the presence of noise and artifacts. This complexity is heightened by the requirement of specialized expertise in correctly identifying various minerals. As the volume of data increases, the task of producing precise segmentation masks becomes even more daunting. Moreover, the variability in rock types, such as sandstone, carbonate, and shale, further complicates the manual segmentation process, making it a labour-intensive and challenging task. Additionally, traditional segmentation models often fail to generalize effectively to new or unseen object classes, as they lack the specialized knowledge needed to recognize and segment such objects accurately [28].

With the progress in foundation models [29], zero-shot segmentation of specific areas is becoming more accessible. SAM employs a vision transformer-based image encoder for feature extraction from images and utilizes prompt encoders to integrate user interactions. This is followed by a mask decoder, which is responsible for producing segmentation outcomes and confidence scores. These results are derived based on the combination of image embedding, prompt embedding, and the output token [19].

The present research is narrowly concentrated on training the SAM model using sandstone, with a focus on binary segmentation. Future endeavours ought to broaden this scope to encompass a diverse array of rock types, including but not limited to carbonate and shale. Additionally, there is a need to expand the study to various digital rock image formats, such as FIB-SEM and thin-section images. The overarching goal is to cultivate a SAM model that is finely tuned and universally applicable for digital rock analysis.

Regarding the SAM model, it is available in three distinct variants: base, large, and huge (ViT-*B*, ViT-*L*, and ViT-*H*) [24]. The 'huge' version is a robust 32-block vision transformer, equipped with approximately 636 million parameters. While ViT-*H* demands more time to generate predictions, the quality of the masks it produces is significantly superior to those of its smaller counterparts (ViT-*B*) [24]. In our study, we employed the smallest image encoder (ViT-*B*) to validate the effectiveness of the fine-tuned SAM model in analysing digital rock images. Additionally, we opted not to fine-tune the image encoder, aiming to minimize computational load. Future research could enhance the model's capacity and its ability to learn more intricate features of rock images by using larger backbone models (such as ViT-*L* or ViT-*H*) and by fine-tuning the image encoder [19]. In addition, a larger training set should be chosen to further enhance the RockSAM performance because in this study, we only adopted the 400 images of sandstone with the size of 1000×1000 and 59 SEM images, which is far smaller compared to the pre-trained SAM's training datasets.

In this study, we have successfully validated the effectiveness of the RockSAM model for binary segmentation. Looking ahead, future research will focus on expanding the capabilities of the fine-tuned SAM model for digital rock images, delving into more complex applications such as multi-phase segmentation and fracture segmentation. Furthermore, there is potential for integrating Grounding DINO with the Segment Anything model (SAM) to enable text input-based segmentation, opening new avenues for enhanced digital rock image segmentation.

## 5. Conclusion

The Segment anything model (SAM) is an advanced segmentation system known for its remarkable zero-shot generalization capabilities, which allow it to segment unfamiliar objects in natural images without additional training. However, its performance is less effective with digital rock images, which are typically complex, have low contrast, and contain small features. To overcome this, the SAM model was fine-tuned specifically for digital rock images. This specialized adaptation has significantly improved its ability to generate accurate segmentation masks for these challenging images, making it a powerful tool for digital rock image analysis.

Notably, the fine-tuned SAM model (RockSAM) maintains its capability to generate segmentation masks without the need for training, making it suitable for use on resource-constrained devices. This aspect is particularly beneficial in situations where computational resources are limited. And it is also anticipating the limitations of current fine-tuned SAM could be overcome by leveraging larger models and increasing the dataset size.

To the best of the authors' knowledge, we are the pioneers in fine-tuning the SAM model for digital rock image segmentation and following-on downstream tasks. By doing so, we have addressed a significant challenge in the field: the time-consuming, labour-intensive, expert-requiring, and expensive process of segmenting digital rock images. Once trained, the model autonomously generates binary segmentation masks, streamlining the segmentation process and potentially leading to more efficient and cost-effective image analysis in this domain.

## Declaration of competing interest

The authors declare that they have no known competing financial interests or personal relationships that could have appeared to influence the work reported in this paper.

## Data availability

Data will be made available on request. The code will upload to the GitHub when the manuscript is accepted.

## Acknowledgement:

This work signifies a collaborative endeavor led by Prof. Shuyu Sun and Prof. Bicheng Yan, generously supported by Saudi Aramco (Dr. Hyung Tae Kwak). We would like to acknowledge the Supercomputing Laboratory at King Abdullah University of Science & Technology (KAUST) in Thuwal, Saudi Arabia, for providing the computational resources utilized in this research. We thank anonymous reviewers for their specific comments and instructive suggestions. Thank the Digital Rocks Portal (https://www.digitalrocksportal.org/) for providing the open source data.